# A Survey of Multibiometric Systems

Harbi AlMahafzah
P.E.T. Research Center,
University Of Mysore,
Mandya, India

hmahafzah@hotmail.com

Ma'en Zaid AlRwashdeh
Department of Study in Computer science,
University Of Mysore,
Mysore, India

Maen_zaid@hotmail.com

## ABSTRACT

Most biometric systems deployed in real-world applications are unimodal. Using unimodal biometric systems have to contend with a variety of problems such as: Noise in sensed data; Intra-class variations; Inter-class similarities; Non-universality; Spoof attacks. These problems have addressed by using multibiometric systems, which expected to be more reliable due to the presence of multiple, independent pieces of evidence.

## General Terms
Biometric.

## 1. INTRODUCTION
The need for reliable user authentication techniques has increased in the wake of heightened concerns about security and rapid advancements in networking, communication and mobility.

A wide variety of applications require reliable verification schemes to confirm the identity of an individual requesting their service. Examples of such applications include; secure access to buildings, computer systems, laptops, cellular phones and ATMs. In the absence of robust verification schemes, these systems are vulnerable to the wiles of an impostor [1][4][5]. Traditionally, passwords (knowledge-based security) and ID cards (token-based security) have been used to restrict access to applications. However, security can be easily breached in these applications when a password is divulged to an unauthorized user or a badge is stolen by an impostor. The emergence of biometrics has addressed the problems that plague traditional verification methods.

## 2. RELATED WORKS
Table-1 shows some recent study on different types of multibiometric with different levels of fusion and the fusion strategies.

## 3. BIOMETERIC SYSTEMS
The term biometric comes from the Greek words bios (life) and metrikos (measure)[5]. Biometric refers to the automatic recognition of individuals based on their physiological and behavioral characteristics. Biometrics systems are commonly classified into two categories: physiological biometrics and behavioral biometrics. Physiological biometrics (fingerprint, iris, retina, hand geometry, face, etc) use measurements from the human body. Behavioral biometrics (signature, keystrokes, voice, etc) use dynamics measurements based on human actions [1][3][8]. These systems are based on pattern recognition methodology, which follows the acquisition of the biometric data by building a biometric feature set, and comparing versus a pre-stored template pattern.

### 3.1 Generic Biometric System
A simple biometric system has a sensor module, a feature extraction module and a matching module (Figure 1). Sensor module (Image acquisition): a suitable sensor to acquire the raw biometric data of an individual to be stored in the database. Feature extraction: a suitable algorithm for feature extraction. It may also require enhancement algorithm to improve the quality of acquired image. Database module: which acts as a respiratory of biometric information? People have to enroll before they can use biometric systems. Enrolment involves a copy of a person's biometric feature being taken, converted into a digital format and stored on an electronic database. Matching module: The extracted features are compared against the stored templates to generate match score. The performance of a biometric system is largely affected by the reliability of the sensor used and the degrees of freedom offered by the features extracted from the sensed signal [3].

Figure-2 shows an example of a biometric system of a fingerprint character. The matching process involves comparing the two-dimensional minutiae patterns extracted from the user's print with those in the template. Figure-2 shows an example of a biometric system of a fingerprint character. The matching process involves comparing the two-dimensional minutiae patterns extracted from the user's print with those in the template.

## 4. CHARACTERISTICS OF BIOMETRIC
Following are the characteristics of biometric:[3][13]

- **Universality:** Every person should have the biometric characteristic.
- **Uniqueness:** No two persons should be the same in terms of the biometric characteristic.
- **Permanence:** The biometric characteristic should be invariant over time.
- **Collectability**: The biometric characteristic should be measurable with some (practical) sensing device.
- **Acceptability:** The particular user population and the public in general should have no (strong) objections to the measuring/collection of the biometric characteristic.
- **Performance:** Refers to the level of accuracy and speed of recognition of the system given the operational and environmental factors involved.
- **Resistance to Circumvention:** Refers to the degree of difficulty required to defeat or bypass the system.

**Table-1: Some Recent Work on Multibiometric**

| Modality | Level of Fusion | Fusion Strategies | Authors |
|---|---|---|---|
| Palmprint and Face | Matching Level | Sum of Score | Nageshkumar, et al [14] |
| Fingerprint and Hand-Geometry | Combination Approach | Sum, Max, Min Scores | Anil Jain, et al [6] |
| Face and Speech | Matching Level | Voting k-NN | A. Teoh, et al [1] |
| Fingerprint, Palmprint, and Hand-Geometry | Feature Level | ANN | Farhat Anwar, et al [7] |
| Speech, Signature, and Face | Macthing Level | Likelihoods Ratio | Yannis Stylianou, et al[15] |
| Audio and Visual Expert (Lipreading) | Decision Level | Optimal Weight (SVM) | Dzati A., et al [16] |
| Face and Fingerprint | Matching Level | Sum, Min-Max, and Zscore | Robert Snelick, et al [11] |
| Fingerprint and Face | Score and Decision | Sum Rule and Likelihoods | Kalyan, et al [17] |
| Face, Fingerprint, and Hand-Geometry | Matching Level | Sum Rule | Arun Ross and Anil Jain [18] |
| Left and Right Iris | Matching Level | Simple Sum | Arun Ross, et al |

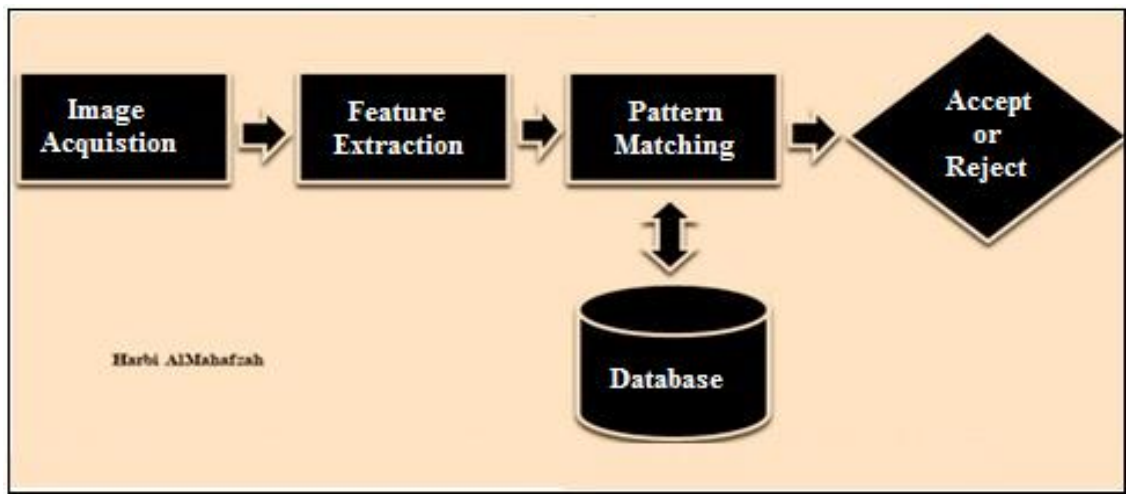

**Figure-1: A Generic Biometric System**

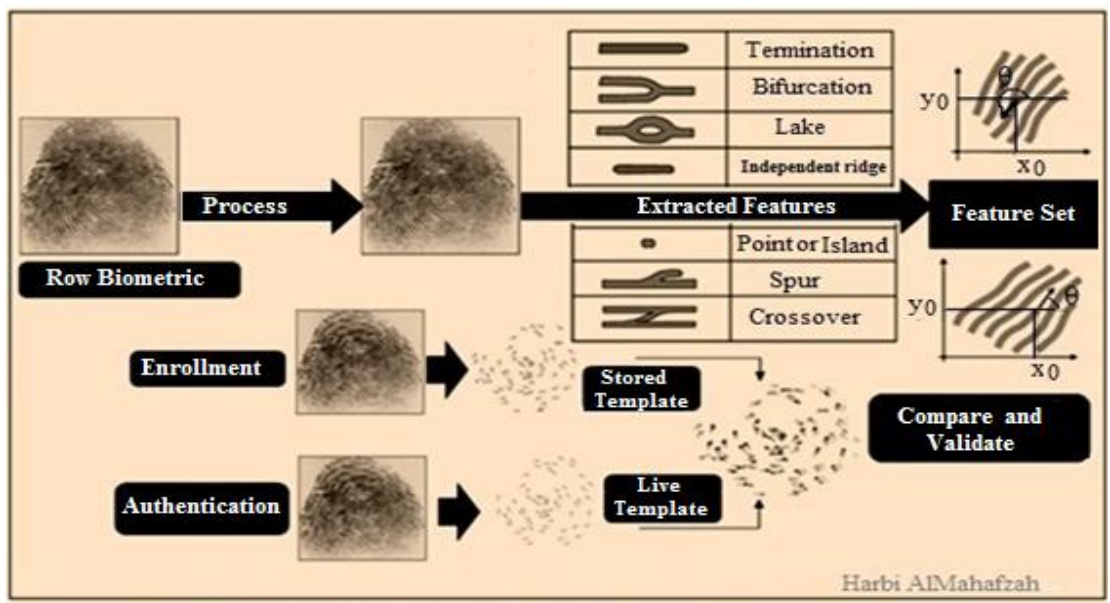

**Figure-2: Example of Biometric System (fingerprint minutiae)**

# 5. POPULAR BIOMETRICS TRAIT (Modalities)

Biometrics systems are commonly classified into two categories: physiological biometrics and behavioral as shown in Figure-3.

## 5.1 Physical Characteristics [5]

- **Face:**

Facial attributes are probably the most common biometric features used by humans to recognize one another. The most popular approaches to face recognition are based on either (i) The location and shape of facial attributes, such as the eyes, eyebrows, nose, lips, and chin and their spatial relationships, or (ii) the overall (global) analysis of the face image that represents a face as a weighted combination of a number of canonical faces.

- **Fingerprint:**

Humans have used fingerprints for personal identification for many decades. A fingerprint is the pattern of ridges and valleys on the surface of a fingertip. It has been empirically determined that the fingerprints of identical twins are different The feature values typically correspond to the position and orientation of certain critical points known as minutiae points.

- **Iris:**

The iris is the annular region of the eye bounded by the pupil and the sclera (white of the eye) on either side. The complex iris texture carries very distinctive information useful for personal recognition of high accuracy and speed. Each Iris is believed to be distinctive. It is possible to detect artificial irises (contact lenses).

- **Palmprint:**

The palms of the human hands contain pattern of ridges and valleys much like the fingerprints. Human palms also contain additional distinctive features such as principal lines and wrinkles that can. It is easy to be captured even with a lower resolution scanner.

- **Retina:**

Retinal recognition creates an "eye signature" from the vascular configuration of the retina which is supposed to be a characteristic of each individual and each eye, respectively. Since it is protected in an eye itself, and since it is not easy to change or replicate the retinal vasculature, this is one of the most secure biometric. Image acquisition requires a person to look through a lens at an alignment target; therefore it implies cooperation of the subject.

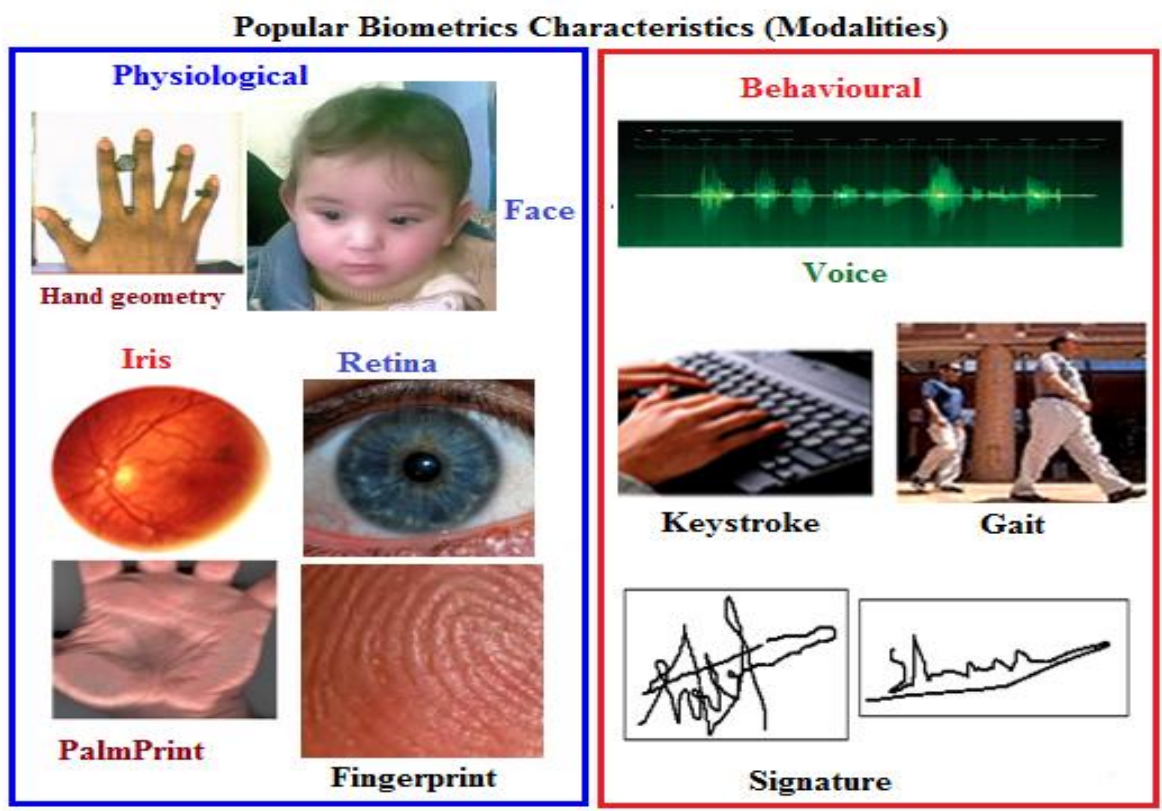

**Figure-3: Some Popular Biometrics Trait.**

## 5.2 Behavioral Characteristics [5]

- **Signature:**

The way a person signs his or her name is known to be characteristic of that individual. In addition to the general shape of the signed name, a signature recognition system can also measure pressure and velocity of the point of the stylus across the sensor pad.

- **Voice:**

Voice is a combination of physical and behavioral biometric characteristics. The physical features of an individual's voice are based on the shape and size of the, vocal tracts, mouth, nasal cavities, and lips that are used in the synthesis of the sound. Feature extraction typically measures formants or sound characteristics unique to each person's vocal tract.

- **Keystroke:**

It is believed that each person types on a keyboard in a characteristic way. This biometric is not expected to be unique to each individual but sufficiently different that permits identity verification.

- **Gait:**

Gait refers to the manner in which a person walks, and is one of the few biometric traits that can be used to recognize people at a distance. Therefore, this trait is very appropriate in surveillance scenarios where the identity of an individual can be surreptitiously established.

# 6. FUNCTIONALITIES OF A BIOMETRIC SYSTEM

biometric system may operate either in the Verification or Identification modes [3][7]. But people have to enroll before they can use biometric systems. Enrolment involves a copy of a person's biometric feature being taken, converted into a digital format and stored on an electronic database as shown in Figure-4.

- **Verification:** an attempt is made to verify the claimed identity of unknown individual. In this mode; biometric system performs a one-to-one comparison of a submitted biometric characteristic (sample) set against a specified stored biometric references, and returns the comparison score and decision. "Is this person who he claims to be?" as shown in Figure-5.

- **Identification:** an attempt is made to establish the identity of an individual. In this mode; biometric system performs a one-to-many comparison/search process in which a biometric characteristic set against all or part of the database to find biometric references with a specified degree of similarity. "Who is this person?" as shown in Figure-6.

# 7. TYPES OF BIOMETRICS

There are two types of biometrics; Unimodal and Multi-Biometrics.

- **Unimodal:** The unimodal rely on the evidence of a single source of information for authentication (e.g., single fingerprint, face) [2][9]. These systems have to contend with a variety of problems such as [2][3][4][10]: (i) Noise in sensed data; a fingerprint image with a scar or a voice sample altered by cold are examples of noisy data. Noisy data could also result from defective or improperly maintained sensors (e.g., accumulation of dirt on a fingerprint sensor). (ii) Intra-class variations; these variations are typically caused by a user who is incorrectly interacting with the sensor (e.g., incorrect facial pose). (iii) Inter-class similarities; in a biometric system comprising of a large number of users, there may be inter-class similarities (overlap) in the feature space of multiple users. (iv) Non-universality; the biometric system may not be able to acquire meaningful biometric data from a subset of users. For example, fingerprint biometric system, may extract incorrect minutiae features from the fingerprints of certain individuals, due to the poor quality of the ridges. (v) Spoof attacks; this type of attack is especially relevant when using behavioral characteristics.

- **Multibiometric:** The term multibiometrics denotes the fusion of different types of information [7] (e.g., fingerprint and face of the same person, or fingerprints from two different fingers of a person). Figure-7 shows the different types of multibiometrics.

Multibiometrics has addressed some issue related to unimodal such as [6][9]:
- Non-universality or insufficient population coverage (reduce failure to enroll rate which increase population coverage).
- It becomes increasingly difficult for an impostor to spoof multiple biometric traits of a legitimately enrolled individual.
- Multibiometric systems also effectively address the problem of noisy data (illness affecting voice, scar affecting fingerprint.

Multibiometric systems can offer substantial improvement in the matching accuracy of a biometric system depending upon the information being combined and the fusion methodology adopted [1].

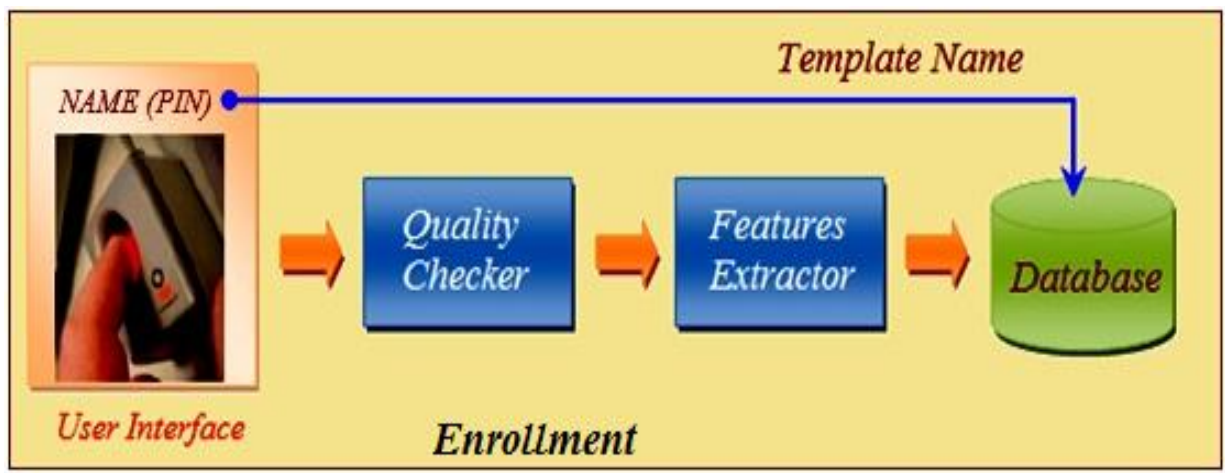

**Figure-4: The Enrollment Process**

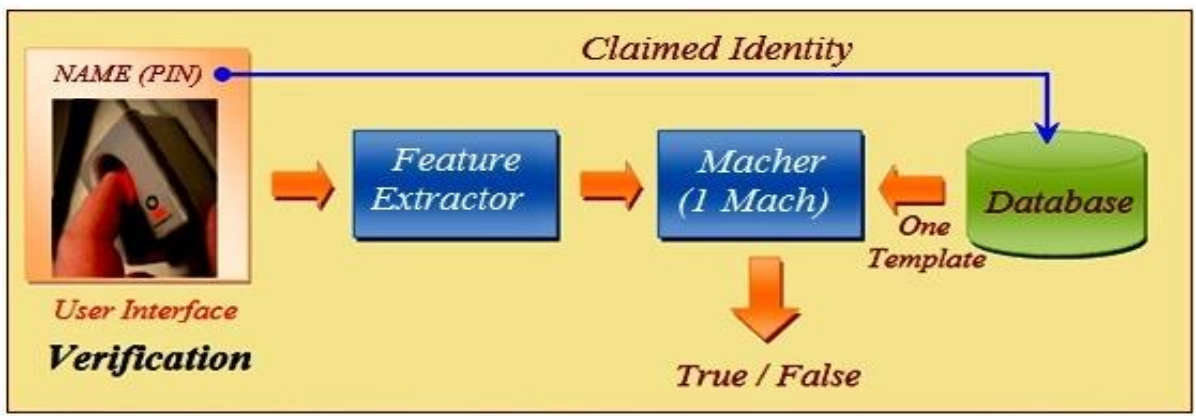

**Figurre-5: The Verification Mode Process**

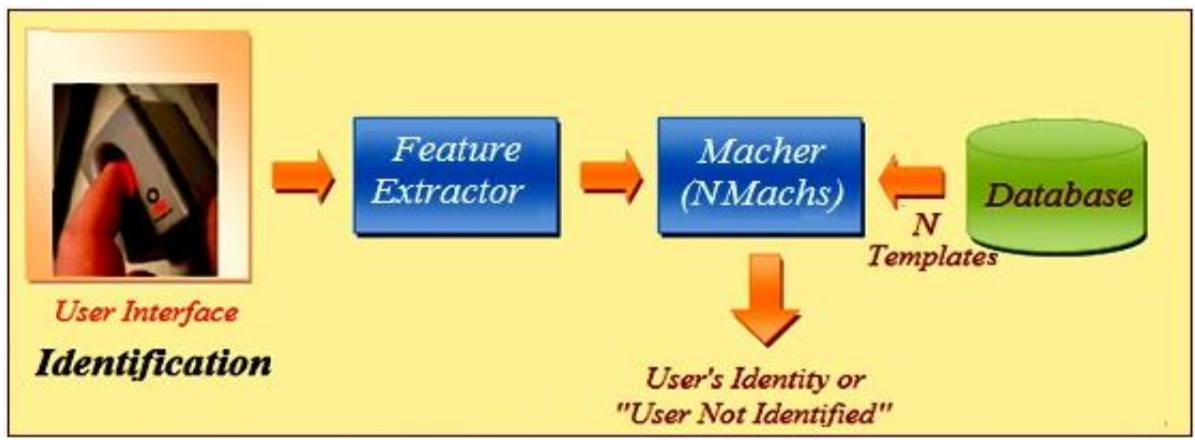

**Figure-6: The Identification Mode Process**

- **Multi sensor:** Multiple sensors can be used to collect the same biometric trait.
- **Multi-modal:** Multiple biometric traits are collected from the same individual, e.g. fingerprint and face, which requires different sensors.
- **Multi-instance:** Multiple units of the same biometric are collected, e.g. fingerprints from two or more fingers.
- **Multi-sample:** Multiple capturing of the same biometric trait are collected during the enrolment and/or authentication phases, e.g. a number of face capturing are taken at different pos and illumination.
- **Multiple algorithms:** Different algorithms for feature extraction and matching are used on the same biometric sample.

## 8. LEVEL OF FUSION IN MULTIBIOMETRIC

One of the most fundamental issues in an information fusion system is to determine the type of information that should be consolidated by the fusion module[12].

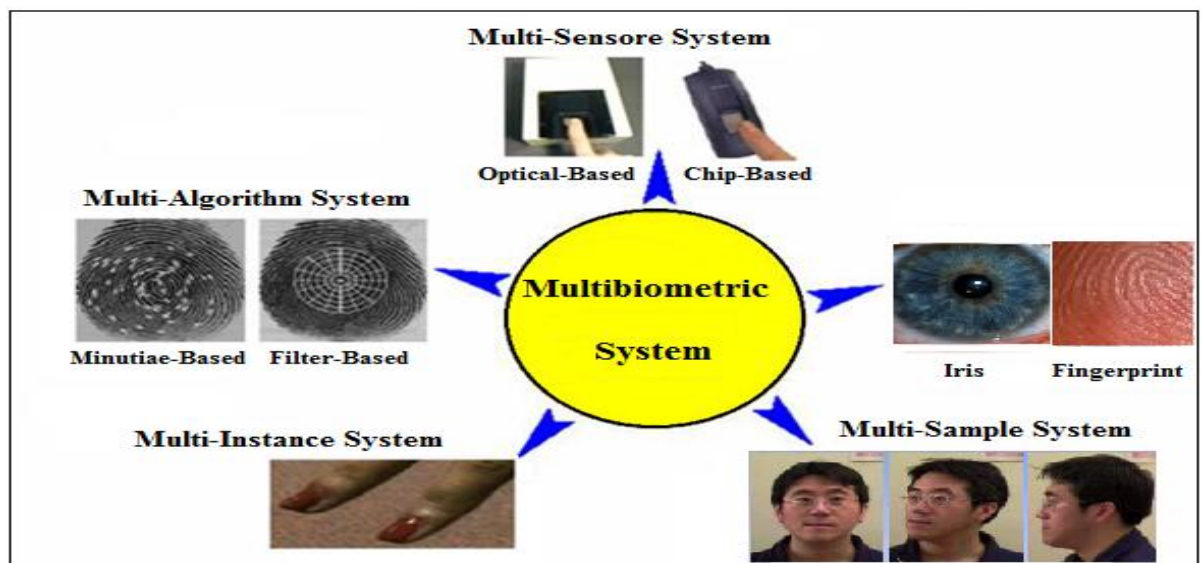

**Figure-7: Different Types of Multibiometric**

- **Sensor level fusion**

Sensor level fusion (Figure-8), entails the consolidation of evidence presented by multiple sources of raw data before they are subjected to feature extraction. Sensor level fusion can benefit multi-sample systems which capture multiple snapshots of the same biometric.

- **Feature Level Fusion**

Feature level fusion (figure-9), consolidating the feature sets obtained from multiple biometric algorithms into a single feature set, after normalization, transformation and reduction schemes.

Feature normalization: The goal of feature normalization is to modify the location (mean) and the scale (variance) of the feature value via a transform function in order to map them into a common domain. (e.g. Min-max normalization, Median normalization)[11].

Feature Selection or Transformation: algorithm use to reduce the dimensionality of the feature set. (e.g. Sequential forward selection, Sequential backward selection, PCA).

- **Score Level Fusion**

In score level fusion (figure-10), the match scores output by nultiple biometric matchers are combined to generate a new match score (a scalar). When match scores output by different biometric matchers are consolidated in order to arrive at a final recognition decision, fusion is said to be done at the match score level. (e.g. similarity score, distance score).

- **Decision Level Fusion**

In a multibiometric system, fusion is carried out at the abstract or decision level (figure-11) when only final decisions are available [3], this is the only available fusion strategy (e.g. AND, OR, Majority Voting, Weighted Majority Voting, Bayesian Decision Fusion).

## 9. INTEGRATION STRATEGIES

The strategy adopted for integration depends on the level at which fusion is performed. Feature selection/reduction techniques may be employed to handle the curse-of-dimensionality problem. Robust and efficient normalization techniques are necessary to transform the scores of multiple matchers into a common domain prior to concatenating them. Different strategies for combining multiple classifiers have been suggested in the literature. Some of the examples of such approaches are: Genetic Algorithm (GA), Ant Colony Optimization (ACO), etc.

## 10. SUMMARY AND CONCLUSIONS

Multibiometric systems alleviate several of the problems present in unimodal systems. By combining multiple sources of information, the multibiometric systems improve matching performance, increase population coverage, deter spoofing, and facilitate indexing. Various fusion levels and scenarios are possible in multibiometric systems, the most popular one being fusion at the matching score level.

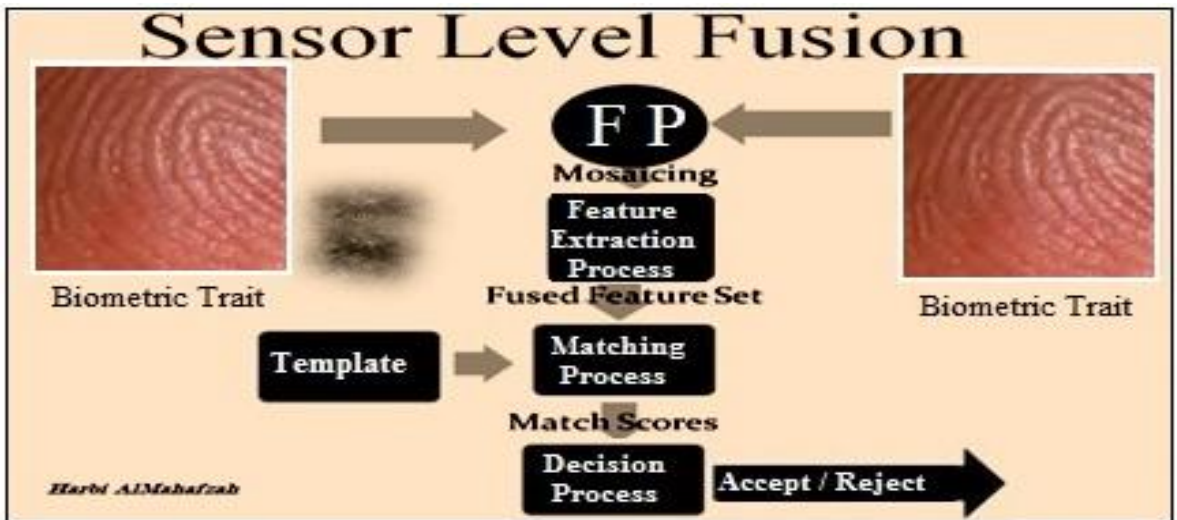

**Figure-8: Sensor Level Fusion**

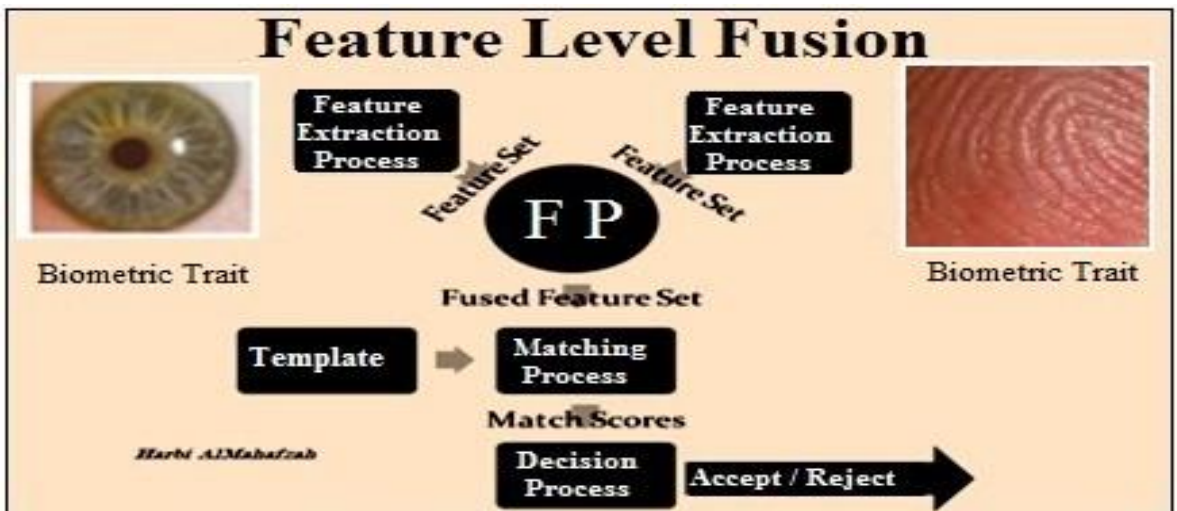

**Figure-9: Feature Level Fusion**

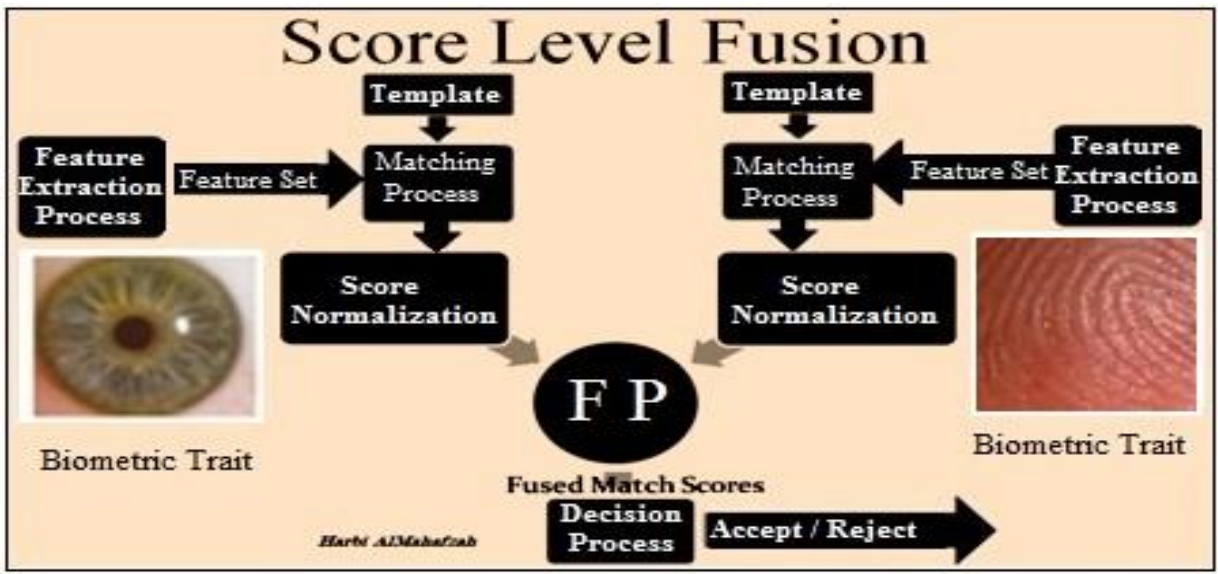

**Figure-10: Score Level Fusion**

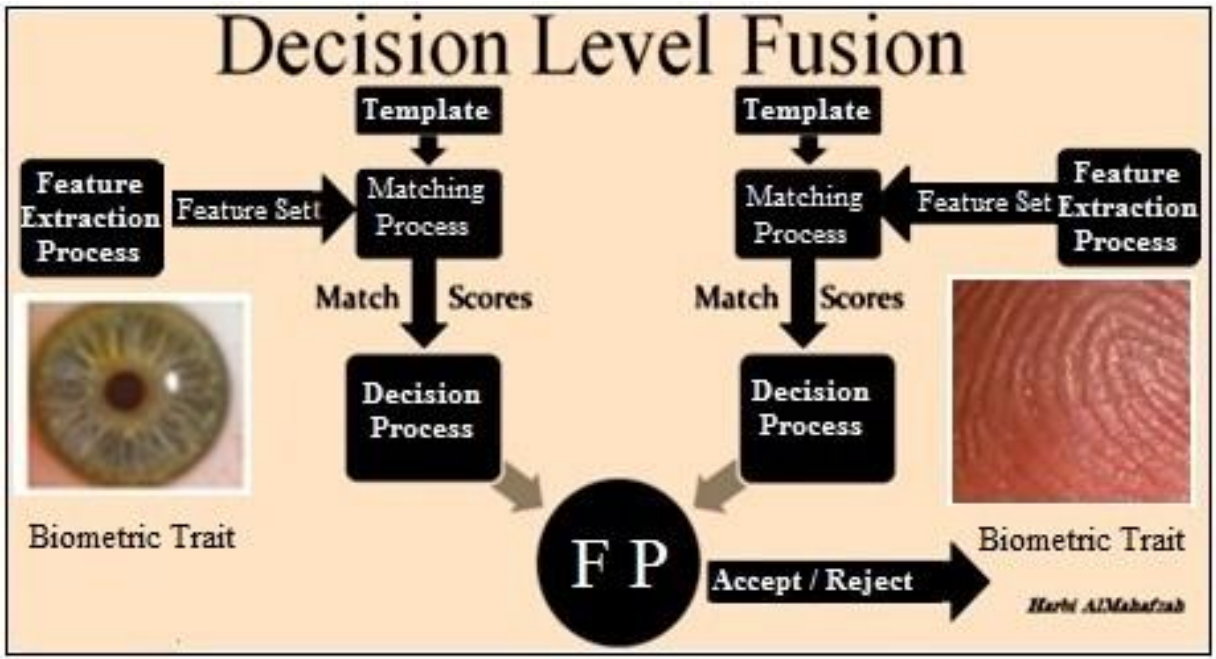

**Figure-11: Decision Level Fusion**